\PassOptionsToPackage{table}{xcolor}
\documentclass[sigconf]{acmart}
\usepackage{booktabs}
\usepackage{pifont}   
\usepackage{makecell} 
\usepackage{colortbl}   
\usepackage{multirow}   
\usepackage{fontawesome5}
\usepackage[most]{tcolorbox}

\AtBeginDocument{%
  }

\setcopyright{acmlicensed}
\copyrightyear{2025}
\acmYear{2025}
\acmDOI{XXXXXXX.XXXXXXX}
\acmConference[Conference acronym 'XX]{Make sure to enter the correct
  conference title from your rights confirmation email}{June 03--05,
  2018}{Woodstock, NY}
\acmISBN{978-1-4503-XXXX-X/2018/06}

\begin{document}

\title[Traits Run Deep:  Psychology-Guided LLM Representations for Personality Assessment]{Traits Run Deep: Enhancing Personality Assessment via Psychology-Guided LLM Representations and Multimodal Apparent Behaviors}

\author{Jia Li}
\authornote{Both authors contributed equally to this research.}
\email{jiali@hfut.edu.cn}
\affiliation{%
  \institution{Hefei University of Technology}
  \city{Hefei}
  \country{China}
}

\author{Yichao He}
\email{heyichao@mail.hfut.edu.cn}
\authornotemark[1]
\affiliation{%
  \institution{Hefei University of Technology}
  \city{Hefei}
  \country{China}
}

\author{Jiacheng Xu}
\email{jcxu@mail.hfut.edu.cn}
\affiliation{%
  \institution{Hefei University of Technology}
  \city{Hefei}
  \country{China}
}

\author{Tianhao Luo}
\email{lth@mail.hfut.edu.cn}
\affiliation{%
  \institution{Hefei University of Technology}
  \city{Hefei}
  \country{China}
}

\author{Zhenzhen Hu}
\authornote{Corresponding author.}
\email{huzhen.ice@gmail.com}
\affiliation{%
  \institution{Hefei University of Technology}
  \city{Hefei}
  \country{China}
}

\author{Richang Hong}
\email{hongrc.hfut@gmail.com}
\affiliation{%
  \institution{Hefei University of Technology}
  \city{Hefei}
  \country{China}
}

\author{Meng Wang}
\email{eric.mengwang@gmail.com}
\affiliation{%
  \institution{Hefei University of Technology}
  \city{Hefei}
  \country{China}
}

\renewcommand{\shortauthors}{Jia Li et al.}

\begin{abstract}
Accurate and reliable personality assessment plays a vital role in many fields, such as emotional intelligence, mental health diagnostics, and personalized education. Unlike fleeting emotions, personality traits are stable, often subconsciously leaked through language, facial expressions, and body behaviors, with asynchronous patterns across modalities. It was hard to model personality semantics with traditional superficial features and seemed impossible to achieve effective cross-modal understanding. To address these challenges, we propose a novel personality assessment framework called \textit{\textbf{Traits Run Deep}}. It employs \textit{\textbf{psychology-informed prompts}} to elicit high-level personality-relevant semantic representations. Besides, it devises a \textit{\textbf{Text-Centric Trait Fusion Network}} that anchors rich text semantics to align and integrate asynchronous signals from other modalities. To be specific, such fusion module includes a Chunk-Wise Projector to decrease dimensionality, a Cross-Modal Connector and a Text Feature Enhancer for effective modality fusion and an ensemble regression head to improve generalization in data-scarce situations. To our knowledge, we are the first to apply personality-specific prompts to guide large language models (LLMs) in extracting personality-aware semantics for improved representation quality. Furthermore, extracting and fusing audio-visual apparent behavior features further improves the accuracy. Experimental results on the AVI validation set have demonstrated the effectiveness of the proposed components, i.e., approximately a 45\% reduction in mean squared error (MSE).  Final evaluations on the test set of the AVI Challenge 2025 confirm our method’s superiority, ranking first in the Personality Assessment track. The source code will be made available at \url{https://github.com/MSA-LMC/TraitsRunDeep}.


\end{abstract}



\received{20 February 2007}
\received[revised]{12 March 2009}
\received[accepted]{5 June 2009}

\maketitle

\section{Introduction}
In daily work settings, personality serves as a key indicator for assessing an individual's behavioral style and their compatibility with specific job roles. It not only influences communication patterns, teamwork, and work efficiency, but also plays a vital role in recruitment and career development \cite{roberts2009back,liem2018psychology,caldwell1998personality}. With the rapid advancement of deep learning and experimental phycology, personality assessment has shifted from traditional questionnaire-based methods to more natural automated approaches, emerging as a critical challenge and opportunity in real-world applications such as affective computing and asynchronous video interviews \cite{cai2024mdpe,liao2024open,mujtaba2021multi}.

To advance this field, a series of personality recognition challenges have been launched.  ECCV 2016 hosted the first global competition on personality recognition from short videos \cite{ponce2016chalearn}, followed by the CVPR 2017 challenge on trait-based job screening \cite{8999746}, driving progress in multimodal modeling of apparent personality. To further advance research in this field, the AVI 2025 competition has organized a Personality Assessment Track. Participants are asked to assess personality traits based on subjects' responses to questions specifically designed by psychologists to elicit corresponding inner traits. 

Existing personality assessment methods primarily rely on feature engineering, where features are extracted using pre-trained models and then processed through deep neural networks \cite{8031881, ghassemi2023unsupervised,koutsoumpis2024beyond,zhang2024can}. However, these approaches often overlook the significant challenges posed by the latent nature of personality, which requires capturing deeper semantic information during feature extraction. Moreover, in personality assessment tasks, existing methods typically concatenate the extracted features in a simple manner and treat all modalities equally, without fully exploring the complex interactions among different features and modalities.

To address these limitations and challenges, we propose a novel framework named \textbf{Traits Run Deep}, which integrates psychology-informed prompts with a Text-Centric Trait Fusion Network for robust personality assessment. For the \textit{psychology-informed prompts}, this work is the first in the field of personality assessment to introduce specific prompts that guide large language models (LLMs) to focus on personality-relevant semantics, thereby generating more accurate personality representations. For the \textit{Text-Centric Trait Fusion Network}, it consists of a Chunk-Wise Projector (CWP), a Cross-Modal Connector (CMC), a Text-Feature Enhancer (TFE), and an ensemble regression head. The CWP is designed to parallelly reduce the dimensionality of high-dimensional features generated by LLMs, alleviating the curse of dimensionality under small-sample conditions. The CMC and TFE aim to maximally capture personality-relevant semantics from different modalities. The ensemble regression head is intended to produce more robust predictions under limited data settings through ensemble learning.

Our approach was evaluated on both the validation and test sets of the AVI Challenge 2025. On the validation set, it achieved an average best MSE of 0.1003 across the four personality dimensions, significantly outperforming the competition baseline of 0.1796. On the test set, we obtained an MSE of 0.12284.


\begin{figure}[t]  %
  \centering
  \includegraphics[width=\linewidth]{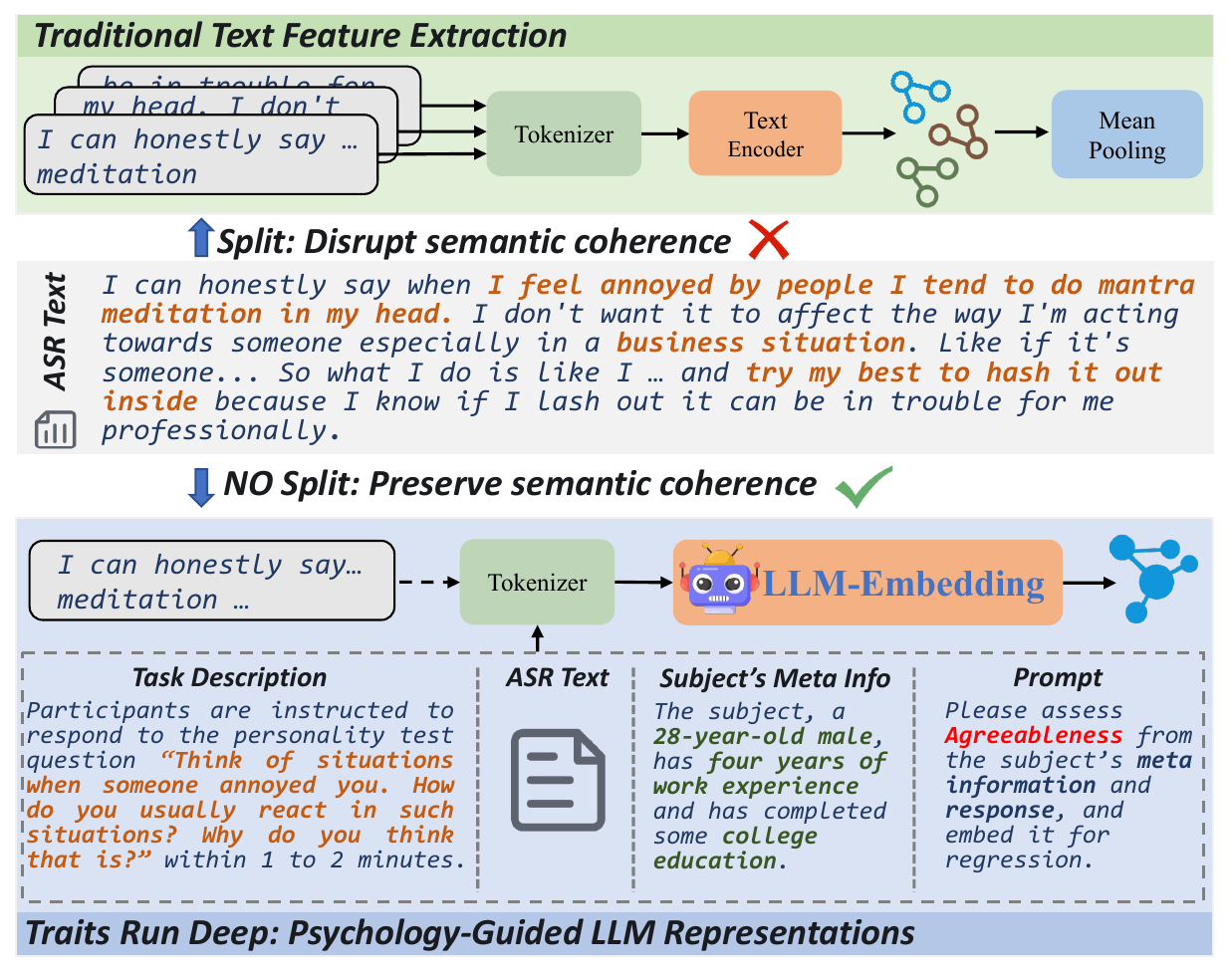}
  \caption{A Comparative Analysis of Traditional Text Feature Extraction and Traits Run Deep's Psychology-Guided LLM Representations.}
  \Description{compare_traits_run_deep_and_traditional method}
  \label{fig:crop_traits_run_deep}
\end{figure}

\begin{figure*}[t]  
  \centering
  \includegraphics[width=\textwidth]{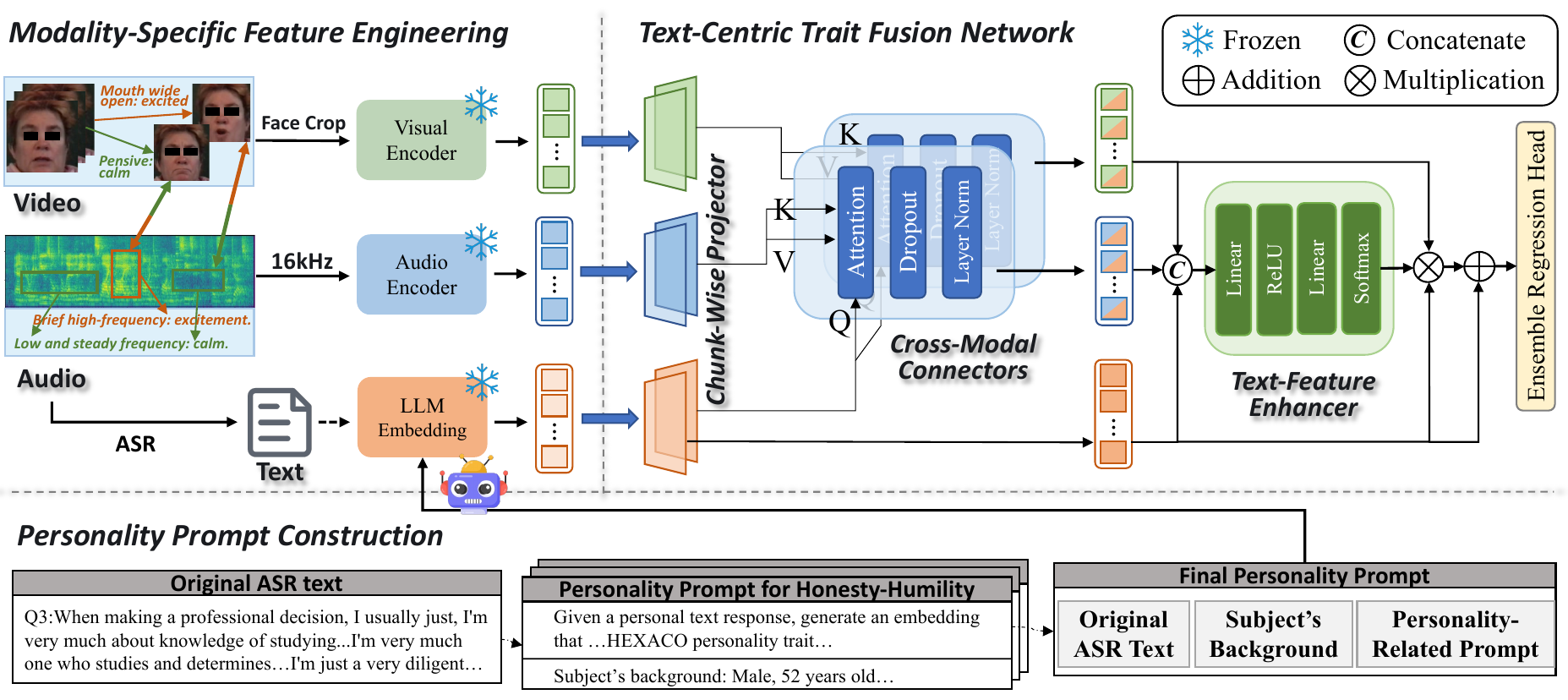}
  \caption{The architecture of our Traits Run Deep framework consists of two main components: (1) Modality-Specific Feature Engineering, which utilizes pre-trained encoders and instruction-following large language models (LLMs) with personality-specific prompts to extract trait-relevant semantic features; and (2) Text-Centric Trait Fusion Network, which incorporates Chunk-wise Projectors, Cross-Modal Connectors, and a Text-Feature Enhancer, carefully designed to align and integrate high-dimensional features from text, audio, and visual modalities.}
  \label{fig:Traits_Run_Deep_framework}
  \Description{Traits Run Deep framework R-Architecture.}
\end{figure*}

\section{Related Works}
\label{sec:related}
\textbf{Automatic Personality Assessment.} Traditional methods for assessing personality are largely based on standardized psychological questionnaires, including NEO-PI-R \cite{costa2008revised}, BFI \cite{fossati2011big}, BFI-2 \cite{soto2017next}, HEXACO-PI-R \cite{ashton2009hexaco}, which often suffers from subjectivity, duplicity and inefficiency. In recent years, deep learning methods have provided a novel direction for personality assessment. Deep learning models can automatically extract informative features from multiple modalities and predict personality traits efficiently and accurately, a process commonly referred to as automatic personality assessment \cite{kedar2015automatic}. Representative datasets for automatic personality assessment include First Impressions \cite{escalante2020modeling} and UDIVA \cite{palmero2021context}. In terms of assessment methods, the study in \cite{liao2024open} establishes an open-source benchmark of deep learning models for personality assessment, evaluating a range of unimodal and multimodal architectures on both the First Impressions and UDIVA datasets.

\textbf{Multimodal Fusion and Modeling.} Multimodal data provide a wealth of personality-related information, encompassing both handcrafted and learned features \cite{junior2019first, vinciarelli2014survey}. Handcrafted audio descriptors such as energy \cite{biel2012youtube}, pause rate \cite{liu2020speech} and Log‑Mel spectrograms \cite{suman2022multi}, visual cues including gaze, pose and motion \cite{biel2012youtube}, and linguistic markers like LIWC and MRC \cite{mairesse2007using} offer interpretability but often miss subtle, implicit signals. Deep learning models bridge this gap by capturing fine-grained representations across modalities. In the visual domain, CNN, CNN‑LSTM, ResNet and ViT can extract detailed features \cite{wei2017deep,subramaniam2016bi,guccluturk2016deep,dosovitskiy2020image}, in the audio domain VGGish and ResNet can encode richer characteristics \cite{suman2022multi,guccluturk2016deep}, and in the textual domain BERT-based approaches can generate context-aware embeddings \cite{tsani2023personality}. Effective fusion of these heterogeneous features is typically achieved through attention mechanisms such as self-attention \cite{vaswani2017attention} and cross-modal attention \cite{tan2019lxmert}. Building on this foundation, \cite{lv2023automated} introduced a topic-guided window-consistency fusion network and \cite{hemamou2021multimodal} developed a hierarchical fusion approach using Gated Multimodal Units at regular intervals. Inspired by these advances, we develop Cross‑Modal Connectors and a Text‑Feature Enhancer to further strengthen multimodal personality representation learning.

\section{Approach}
\label{sec:method}

Our research focuses on estimating four dimensions of the HEXACO personality model, including Honesty-Humility, Extraversion, Agreeableness, and Conscientiousness. To enhance personality representation and improve model performance, we propose the \textbf{Traits Run Deep} framework, which consists of \textit{Modality-Specific Feature Engineering} and a cross-modal interaction network called the \textit{Text-Centric Trait Fusion Network}. In the textual modality, psychology-informed prompts are employed to elicit high-level personality-relevant semantic representations, enabling more effective personality prediction.

\subsection{Modality-Specific Feature Engineering}
\subsubsection{\textbf{Visual-Audio Feature Extraction}}
To ensure consistency across samples, we first extract 16kHz audio tracks from the original video data using FFmpeg\footnote{\url{https://ffmpeg.org}}. The extracted audio is then transcribed into text using the Whisper-small model \cite{radford2023robust}. Meanwhile, facial regions are detected and cropped from video frames using Arc2Face \cite{papantoniou2024arc2face}.

For audio features, we use the pre-trained Emotion2Vec \cite{ma2023emotion2vec} model, which captures emotional and speaker-specific speech patterns. Unlike traditional encoders, it is trained with emotional supervision and self-supervised tasks, preserving affective and prosodic cues linked to personality traits. For video features, cropped face images are input to SigLIP2 \cite{tschannen2025siglip}, a vision-language pre-trained model, to extract high-level embeddings of facial expressions.

In practice, all the pre-trained encoders are frozen during feature extraction, enabling us to leverage their generalizable representations while minimizing computational cost.

\subsubsection{\textbf{Text Feature Extraction}}
Traditional methods, such as BERT-based models for text feature extraction, have difficulty capturing deep personality-related information, as demonstrated in our ablation study (Section~\ref{sec:experiments}). This limitation stems from both the dataset characteristics and the feature extractor’s capacity. Transcribed speech data often contain rich personality cues resulting in long text sequences. Due to tokenizer token limits, traditional models apply sliding-window segmentation, which disrupts textual coherence and hinders personality modeling. Additionally, the pre-training corpora of these models are often limited in size and diversity, causing them to extract only shallow, general features and fail to capture the nuanced semantic information essential for personality inference.

To extract high-level and personality-relevant text features, we employ SFR-Embedding-Mistral \cite{caspari2024beyond}, a large language model based on Mistral-7B \cite{jiang2023mistral7b} and E5-mistral-7b-instruct \cite{wang2023improving}. This model is trained on extensive corpora, endowing it with rich prior knowledge crucial for effectively capturing personality-related representations. Furthermore, to leverage the instruction-following capability of large language models, we design psychology-informed prompts tailored to each personality trait. The prompt format is as follows:

\begin{center}
\begin{tcolorbox}[colback=gray!5!white, colframe=black!75!black, width=0.85\linewidth, boxrule=0.8pt, arc=4pt, auto outer arc, enhanced]
\centering
\textbf{Prompt =} \textless{} \textit{Personality Task Description} \textgreater{} + \textless{} \textit{ASR Text} \textgreater{} + \textless{} \textit{Subject’s Meta Information} \textgreater{}
\end{tcolorbox}
\end{center}

As illustrated in Figure~\ref{fig:crop_traits_run_deep}, the personality task description guides the model toward the current assessment target, while the subject's meta information includes demographic information such as gender, age, education level, and work experience. This is intended to help the LLM integrate the subject's meta information with their linguistic expressions, facilitating the extraction of personality-relevant features.

In practice, we experimented with a variety of personality-related prompts and empirically selected the optimal prompt for each trait based on validation performance. This method aims to maximize the use of LLMs’ prior knowledge about personality, thereby enabling the extraction of more trait-specific text features.

\subsection{Text-Centric Trait Fusion Network}
\subsubsection{\textbf{Chunk-wise Projector}}
Given the limited dataset size, directly using high-dimensional features from large models can lead to the curse of dimensionality. To mitigate this, we design a Chunk-Wise Projector that divides the input feature vector into smaller segments and projects each independently. This approach helps integrate fine-grained semantic information and supports localized feature learning. The detailed mathematical formulation is as follows:

First, the input feature vector $\mathbf{x} \in \mathbb{R}^{D}$ is divided into $N$ non-overlapping chunks along the feature dimension:
\begin{equation}
\mathbf{x} = [\mathbf{x}^{(1)}; \mathbf{x}^{(2)}; \dots; \mathbf{x}^{(N)}]
\label{eq:chunk-split}
\end{equation}

Here, $\mathbf{x}^{(i)} \in \mathbb{R}^{d}$ denotes the $i$-th chunk of dimension $d$, and $d = \frac{D}{N}$, where $D$ is the total dimension of the input vector.

Each chunk $\mathbf{x}^{(i)}$ is then independently projected through a lightweight feedforward network consisting of two linear layers, ReLU activation, LayerNorm, and Dropout, resulting in the transformed chunk $\mathbf{z}^{(i)}$, as shown in Equation~\ref{eq:chunk-ffn}.

\begin{equation}
\mathbf{z}^{(i)} = \mathrm{Dropout}\big(\mathrm{LayerNorm}\big(\mathrm{ReLU}(W_2 \cdot (\mathrm{ReLU}(W_1 \cdot \mathbf{x}^{(i)})))\big)\big)
\label{eq:chunk-ffn}
\end{equation}

In this formulation, $W_1 \in \mathbb{R}^{h \times d}$ and $W_2 \in \mathbb{R}^{d' \times h}$ are learnable weight matrices of the feedforward layers, where $h$ is the hidden dimension and $d'$ is the output dimension per chunk.

Finally, the projected chunks are concatenated to form the final output $\mathbf{z}$, as shown in Equation~\ref{eq:concat}.

\begin{equation}
\mathbf{z} = [\mathbf{z}^{(1)}; \mathbf{z}^{(2)}; \dots; \mathbf{z}^{(N)}] \in \mathbb{R}^{D'}
\label{eq:concat}
\end{equation}

Here, $D' = N \times d'$, where $d'$ is the projected dimension of each chunk.

\subsubsection{\textbf{Cross-Model Connector}}
To enable effective interaction between modalities, we design a Cross-Modal Connector, which allows one modality to attend to another via multi-head attention. Specifically, we use \textbf{text features as queries} and \textbf{audio or visual features as keys and values}, enabling the model to capture audio or visual clues that are relevant to the linguistic content.

Given the input text feature \(\mathbf{x}_{\text{text}} \in \mathbb{R}^{D_t}\) and video feature \(\mathbf{x}_{\text{video}} \in \mathbb{R}^{D_v}\), we first project them into a common multi-head attention space. The query matrix \(\mathbf{Q}\) is obtained by applying a learned linear transformation \(\mathbf{W}_q\) to the text feature, followed by reshaping:
\begin{equation}
\mathbf{Q} = \mathrm{reshape}(\mathbf{W}_q \mathbf{x}_{\text{text}}) \in \mathbb{R}^{H \times d_h}
\label{eq:query}
\end{equation}
Here, $H$ denotes the number of attention heads, and $d_h = \frac{D_o}{H}$ represents the dimension of each head, where $D_o$ is the total output dimension. The key and value matrices, $\mathbf{K}$ and $\mathbf{V}$, are obtained in a similar manner by applying separate learnable linear projections.

Once the query and key matrices are obtained, the scaled dot-product attention weights for each head \(i\) are computed and normalized via Softmax as  
\begin{equation}
\alpha_i = \mathrm{Softmax}\left(\frac{\langle \mathbf{Q}_i, \mathbf{K}_i \rangle}{\sqrt{d_h}}\right)
\label{eq:attention_weights}
\end{equation}
The attended features \(\mathbf{C}_i\) are obtained by weighting the value vectors:  
\begin{equation}
\mathbf{C}_i = \alpha_i \cdot \mathbf{V}_i
\label{eq:attended_features}
\end{equation}
All heads’ outputs are concatenated to form the context vector \(\mathbf{z} \in \mathbb{R}^{D_{\text{out}}}\):  
\begin{equation}
\mathbf{z} = [\mathbf{C}_1; \mathbf{C}_2; \dots; \mathbf{C}_H]
\label{eq:context_vector}
\end{equation}
Finally, the output is produced by applying a linear projection with learnable weights \(\mathbf{W}_o \in \mathbb{R}^{D_{\text{out}} \times D_{\text{out}}}\), followed by dropout and layer normalization:  
\begin{equation}
\mathbf{z}_{\mathrm{out}} = \mathrm{LayerNorm} \big( \mathrm{Dropout}(\mathbf{W}_o \mathbf{z}) \big)
\label{eq:final_output}
\end{equation}

\subsubsection{\textbf{Text-Feature Enhancer}}
To enrich the semantic representation of textual features, we design a Text-Feature Enhancer. Given the input feature triplet \((\mathbf{x}_{at}, \mathbf{x}_{vt}, \mathbf{x}_t) \in \mathbb{R}^{B \times D}\), where \(\mathbf{x}_{at}\) and \(\mathbf{x}_{vt}\) are audio-text and video-text fused features obtained via our Cross-Modal Connector, respectively, and \(\mathbf{x}_t\) denotes the original text feature, we formulate the method as follows:

Each input feature is independently projected into a common latent space through linear layers. This process can be compactly formulated as:
\begin{equation}
\begin{bmatrix}
\hat{\mathbf{x}}_{at} \\
\hat{\mathbf{x}}_{vt} \\
\hat{\mathbf{x}}_t
\end{bmatrix}
=
\begin{bmatrix}
W_{at} & 0 & 0 \\
0 & W_{vt} & 0 \\
0 & 0 & W_t
\end{bmatrix}
\begin{bmatrix}
\mathbf{x}_{at} \\
\mathbf{x}_{vt} \\
\mathbf{x}_t
\end{bmatrix}
\label{eq:modal_projection}
\end{equation}
where \(\mathbf{x}_{at}\), \(\mathbf{x}_{vt}\), and \(\mathbf{x}_t\) denote the input features of audio-text, video-text, and text modalities, respectively; and \(W_{at}\), \(W_{vt}\), and \(W_t\) are the corresponding learnable projection matrices.

Then, a weighted sum of the transformed features is computed according to the dynamically learned gates \( g_1 \), \( g_2 \) and \( g_3 \): 
\begin{equation}
\mathbf{x}_{\text{fused}} = g_1 \cdot \hat{\mathbf{x}}_{at} + g_2 \cdot \hat{\mathbf{x}}_{vt} + g_3 \cdot \hat{\mathbf{x}}_t
\label{eq:fused_features}
\end{equation}

To retain the original textual information, a residual connection is added between the fused output and the raw text feature, followed by dropout, layer normalization, and an optional projection to the output dimensionality:
\begin{equation}
\mathbf{x}_{\text{out}} = \mathrm{LayerNorm}\big(\mathbf{x}_t + \mathrm{Dropout}(\mathbf{x}_{\text{fused}})\big)
\label{eq:enhanced_text_output}
\end{equation}

\subsubsection{\textbf{Ensemble Regression Head}}
Considering the limited size of the dataset, solely relying on a single regression network may lead to unstable predictions. To address this issue, we employ an Ensemble Regression Head composed of 32 independent regression sub-networks to perform ensemble learning. Each sub-network consists of three fully connected layers with ReLU activations, which map the enhanced feature vector to the target prediction space.

The final prediction is computed as the average of all sub-network outputs:
\begin{equation}
\hat{\mathbf{y}} = \frac{1}{32} \sum_{i=1}^{32} \mathrm{MLP}_i(\mathbf{x})
\label{eq:ensemble_prediction}
\end{equation}

\section{Experiments}
\label{sec:experiments}
To evaluate the effectiveness of our Traits Run Deep in HEXACO personality prediction, we conducted experiments on the AVI 2025 dataset, including ablation studies on the Psychology-Guided LLM Representations and the Text-Centric Trait Fusion Network. This section describes the datasets used for training and evaluation, details the experimental setup, reports the ablation results, and highlights the best performance.

\subsection{Dataset}
We conduct experiments on the official AVI Challenge 2025 dataset, which contains structured video interviews from 644 participants. Each interview includes six questions: two general and four related to specific HEXACO personality traits. Ground-truth scores range from 1 to 5 for the personality-related questions are provided for Track 1 . The dataset also includes demographic metadata such as age, gender, education, and work experience. Model performance on Track 1 is measured using Mean Squared Error (MSE).

\subsection{Experimental Setup}
During the feature engineering stage, we used two NVIDIA A800 80GB GPUs to efficiently extract audio and visual features and determine suitable prompts for the four personality traits. All model training and ablation experiments were conducted on a single NVIDIA RTX 4090 24GB GPU.

To optimize the model, we use the Adam optimizer with a learning rate of 1e-4 and a batch size of 32. Initial dropout rates are set to 0.2 for Chunk-Wise Projectors, 0.3 for Cross-Modal Connectors, and 0.1 for the Text-Feature Enhancer. Grid search is employed to select optimal hyperparameters. We also apply Exponential Moving Average and K-fold ensemble strategies during the 200 training epochs.

\subsection{Ablation Study}
\textbf{Ablation on Psychology-Guided LLM Representations and modalities.} By leveraging Psychology-Guided LLM Representations, we propose Traits Run Deep. Table~\ref{tab:personality_mse} presents the performance comparison in four personality dimensions when using different modalities as input. We compare our method against several baselines. LLM-Embedding (using SFR-Embedding-Mistral \cite{caspari2024beyond} in practice) refers to the model that does not incorporate any personality-specific prompts, whereas LLM-Embedding* indicates the variant that utilizes the most effective prompt customized for each personality trait. Among traditional models, Flan-T5-large \cite{chung2024scaling}, DeBERTa-v3-large \cite{he2021debertav3}, and RoBERTa-base \cite{liu2019roberta} perform relatively well, but still exhibit significant fluctuations across different personality dimensions, indicating limited generalization ability. In contrast, our method (LLM-Embedding*) achieves the best performance across all dimensions, demonstrating that personality-relevant prompts can significantly enhance the model’s ability to capture personality-related semantics.

In the audio-only setting, models like Whisper \cite{radford2023robust} and Emotion2Vec \cite{ma2023emotion2vec} show moderate ability to capture personality cues, with Whisper-base achieving the lowest MSE on Honesty-Humility (0.1608) and Emotion2Vec-base excelling on Agreeableness (0.1890). However, audio alone is insufficient to fully model the HEXACO traits. In the video-only setting, ViTMAE \cite{he2022masked} and SigLIP2 \cite{tschannen2025siglip} deliver more balanced results, with SigLIP2 outperforming ViTMAE across all dimensions and achieving the lowest MSE on Emotionality (0.2293) and Honesty-Humility (0.1845). Visual features provide valuable nonverbal information but remain less effective than text in capturing deeper personality semantics. Overall, audio-only or video-only unimodal results are suboptimal, but they can complement the text modality to achieve better performance.

\begin{table}[ht]
\centering
\caption{Mean Squared Error (MSE) on the AVI validation set for predicting four HEXACO personality traits using features from different modalities.}
\label{tab:personality_mse}
\begin{tabular}{lcccc}
\toprule
\textbf{Feature} & \textbf{H} $\downarrow$ & \textbf{E} $\downarrow$ & \textbf{A} $\downarrow$ & \textbf{C} $\downarrow$ \\
\midrule
\rowcolor{gray!15}
\multicolumn{5}{l}{\textit{Traditional Model Features (Audio-Only)}} \\

Whisper-base \cite{radford2023robust}        & \textbf{0.1608} & 0.2497 & \underline{0.2023} & 0.1879 \\
Whisper-large \cite{radford2023robust}    & 0.1862 & \underline{0.2166} & 0.2036 & \textbf{0.1541} \\
Emotion2Vec-base \cite{ma2023emotion2vec}      & \underline{0.1742} & 0.2198 & \textbf{0.1890} & \underline{0.1636} \\
Emotion2Vec+ base \cite{ma2023emotion2vec} & 0.1886 & 0.2431 & 0.2098 & 0.1823 \\
Emotion2Vec+ seed \cite{ma2023emotion2vec} & 0.1892 & \textbf{0.2086} & 0.2179 & 0.1724 \\

\rowcolor{gray!15}
\multicolumn{5}{l}{\textit{Traditional Model Features (Video-Only)}} \\
ViTMAE \cite{he2022masked}       & 0.1852 & 0.2681 & 0.2187 & 0.1852 \\
SigLIP2 \cite{tschannen2025siglip}  & \textbf{0.1845} & \textbf{0.2293} & \textbf{0.2183} & \textbf{0.1811} \\

\rowcolor{gray!15}
\multicolumn{5}{l}{\textit{Traditional LLM-Based Representations (Text-Only)}} \\

ALBERT-base-v2 \cite{lan2019albert}      & 0.1466 & 0.1720 & 0.1308 & 0.1622 \\
ALBERT-large-v2 \cite{lan2019albert}     & 0.1579 & 0.1980 & 0.1340 & 0.1626 \\
ALBERT-xxlarge-v2 \cite{lan2019albert}    & 0.1456 & 0.1702 & 0.1330 & 0.1525 \\
BERT-base \cite{devlin2019bert}     & 0.1562 & 0.2043 & 0.1502 & 0.1509 \\
BERT-large \cite{devlin2019bert}     & 0.1402 & \underline{0.1613} & 0.1649 & 0.1435 \\
DeBERTa-v3-base \cite{he2021debertav3}     & 0.1551 & 0.1781 & 0.1485 & \textbf{0.1264} \\
DeBERTa-v3-large \cite{he2021debertav3}     & \underline{0.1272} & 0.2035 & 0.1742 & 0.1522 \\
Flan-T5-base \cite{chung2024scaling}       & 0.1474 & 0.1813 & 0.2137 & 0.1717 \\
Flan-T5-large \cite{chung2024scaling}       & \textbf{0.1243} & \textbf{0.1333} & \underline{0.1177} & 0.1309 \\
RoBERTa-base \cite{liu2019roberta}         & 0.1368 & 0.1644 & \textbf{0.1111} & \underline{0.1302} \\
RoBERTa-large \cite{liu2019roberta}        & 0.1498 & 0.1766 & 0.1240 & 0.1463 \\

\rowcolor{gray!15}
\multicolumn{5}{l}{\textit{Recent LLM-Based Representations (Text-Only)}} \\

Vicuna-7B \cite{zheng2023judging}           & 0.1635 & 0.2144 & 0.1760 & 0.1663 \\
LLM-Embedding         & \underline{0.1158} & \underline{0.1551} & \underline{0.1321} & \underline{0.1181} \\
\rowcolor{cyan!15}
\textbf{LLM-Embedding*} & \textbf{0.1095} & \textbf{0.1157} & \textbf{0.0971} & \textbf{0.1052} \\
\midrule
\rowcolor{gray!3}
\textcolor{gray}{\textit{Audio+Video+Text}} & \textcolor{gray}{\textit{0.1072}} & \textcolor{gray}{\textit{0.1003}} & \textcolor{gray}{\textit{0.0981}} & \textcolor{gray}{\textit{0.0957}} \\
\bottomrule
\end{tabular}
\end{table}

\begin{figure}[t]  %
  \centering
  \includegraphics[width=0.88\linewidth]{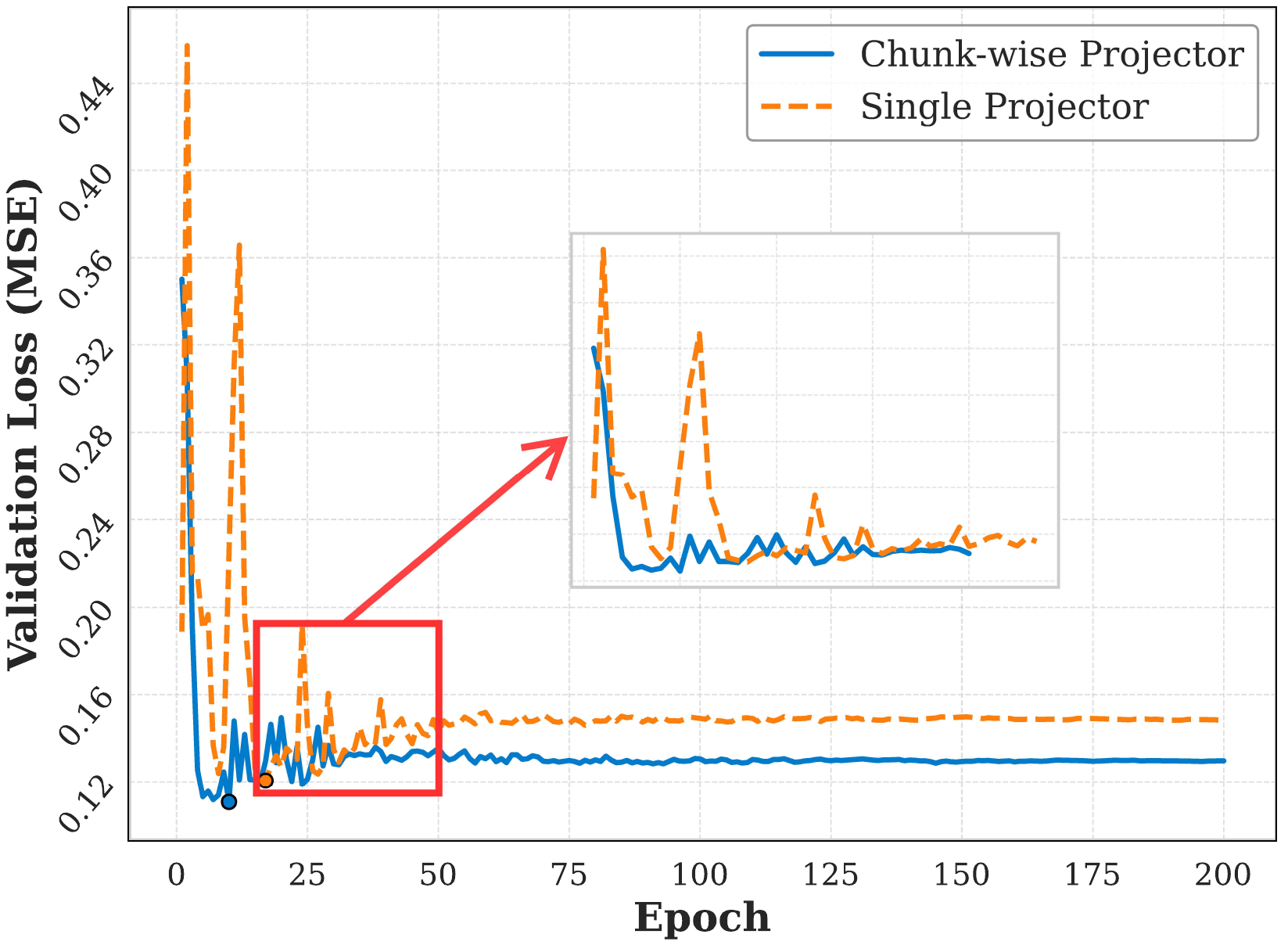}
  \caption{Comparison of validation loss between the Chunk-wise Projector and the Single Projector for the Honesty-Humility trait.}
  \Description{compare_val_loss_chunk_vs_single}
  \label{fig:chunk_vs_single}
\end{figure}

\noindent\textbf{Ablation on Chunk-Wise Projector.} To evaluate the effectiveness of the proposed Chunk-wise Projector, we conducted an ablation study comparing it with the conventional Single Projector. As shown in Figure~\ref{fig:chunk_vs_single}, the Chunk-wise Projector not only demonstrates significantly faster and more stable convergence during training but also achieves superior performance in terms of both minimum and final MSE. In contrast, the Single Projector directly projects high-dimensional inputs and suffers from the "curse of dimensionality," resulting in suboptimal compression and potential loss of crucial modality-specific information. As a result,it exhibits less stable learning behavior and slower convergence, with consistently higher validation loss throughout the training process. These observations validate the superior representation capability of the Chunk-wise strategy in handling high-dimensional multimodal features.

\noindent\textbf{Ablation on Cross-Modal Connectors and Text-Feature Enhancer.}
To effectively integrate personality cues from different modalities, we designed the Cross-Modal Connectors (CMC) and the Text-Feature Enhancer (TFE). Table~\ref{tab:ablation_fusion_strategy} presents the ablation study results across four personality dimensions. The experiments show that simple feature concatenation performs poorly, meaning direct fusion cannot fully capture the relationships and complementary information between modalities. Utilizing only the Cross-Modal Connectors significantly improves performance, reducing the average mean squared error (MSE) to around 0.12. This demonstrates that CMC effectively models the semantic interactions between text and audio/video modalities. Meanwhile, employing only the Text-Feature Enhancer also leads to performance gains (MSE approximately 0.13), suggesting that auxiliary modalities enhance the representation power of the text modality and improve its ability to capture personality traits. Notably, combining both CMC and TFE yields the best results across all personality dimensions, further lowering the MSE to about 0.10. This demonstrates their complementary strengths in capturing complex multimodal personality traits and improving overall personality modeling.

\begin{table}[h]
\centering
\caption{Ablation results on the AVI validation set for Cross-Modal Connectors (CMC) and Text-Feature Enhancer (TFE).}
\label{tab:ablation_fusion_strategy}
\begin{tabular}{cc|ccccc}
\toprule
\textbf{CMC} & \textbf{TFE} & \textbf{H} $\downarrow$ & \textbf{E} $\downarrow$ & \textbf{A} $\downarrow$ & \textbf{C} $\downarrow$ & \textbf{Avg.} $\downarrow$ \\
\midrule
\ding{51} & \ding{55} & 0.1243 & 0.1418 & \underline{0.1192} & 0.1103 & \underline{0.1239} \\
\ding{55} & \ding{51} & \underline{0.1198} & \underline{0.1355} & 0.1736 & \underline{0.1057} & 0.1336 \\
\rowcolor{cyan!15}
\ding{51} & \ding{51} & \textbf{0.1072} & \textbf{0.1003} & \textbf{0.0981} & \textbf{0.0957} & \textbf{0.1003} \\
\midrule
\rowcolor{gray!15}
\multicolumn{2}{c|}{Concatenate} & 0.1981 & 0.2212 & 0.2219 & 0.1883 & 0.2074 \\
\bottomrule
\end{tabular}
\end{table}

\noindent\textbf{Ablation on Ensemble Regression Head. }
To evaluate the effectiveness of the Ensemble Regression Head, we conducted an ablation study by replacing it with a single regression layer. Across five experiments with identical parameter settings, the standard deviation of prediction errors decreased from 0.0096 to 0.0031. This indicates that the ensemble design improves the stability of the model by aggregating multiple regression outputs, which is particularly beneficial under limited training data scenarios.

\subsection{Comparison with Competitors}
Finally, after tuning our model framework to its optimal configuration, we conducted a comprehensive evaluation on the AVI 2025 Track 1 test set. As shown in In Table~\ref{tab:avi2025_final_mse}, our team (HFUT-VisionXL) achieved the lowest average MSE on all four personality traits. These results indicate that our method is effective for multimodal personality assessment and demonstrates reliable generalization ability.
\definecolor{gold}{rgb}{1.0, 0.84, 0}
\definecolor{silver}{rgb}{0.75, 0.75, 0.75}
\definecolor{bronze}{rgb}{0.8, 0.5, 0.2}

\begin{table}[h]
\centering
\caption{Final mean squared error (MSE) results on the AVI2025 Track 1 test set. Our team (\textbf{HFUT-VisionXL}) achieves the best performance.}
\label{tab:avi2025_final_mse}
\begin{tabular}{c l l c}
\toprule
\textbf{Rank} & \textbf{Submitter} & \textbf{Team Name} & \textbf{$\mathbf{MSE}_{\text{Avg.}}$} $\downarrow$ \\
\midrule
\rowcolor{cyan!15}
~\textcolor{gold}{\faTrophy} 1st & \textbf{ArchieHe}     & \textbf{HFUT-VisionXL}     & \textbf{0.12284} \\
~\textcolor{silver}{\faTrophy} 2nd & Jezoid               & —                 & 0.13724 \\
~\textcolor{bronze}{\faTrophy} 3rd & CAS-MAIS             & CAS-MAIS          & 0.14351 \\
4th & l\_wen               & The innovators    & 0.14492 \\
5th & ABC-Lab              & —                 & 0.16770 \\
6th & hdd                  & Winner-Team                 & 0.18909 \\
7th & HSEmotion            & —                 & 0.19731 \\
8th & abhisheksingh        & —                 & 0.19779 \\
9th & nzq                  & DERS              & 0.20612 \\
10th & SonyLai             & DERS              & 0.20674 \\
11th & YouTu\_TX           & USTC-IAT-United   & 0.22914 \\
12th & xtli                & HandX             & 0.23824 \\
13th & wjno1               & —                 & 0.24358 \\
14th & gkdx2               & —                 & 1.89703 \\
\bottomrule
\end{tabular}
\end{table}

\section{Conclusion}
\label{sec:conclusion}
In this paper, we proposed \emph{Traits Run Deep}, a novel framework for multimodal personality assessment that combines psychology-guided LLM representations with audio-visual behavioral features. Our approach addressed a critical challenge in personality computing: the difficulty of capturing latent personality traits from the multimodal clues exhibited by a person while speaking.
By incorporating psychology-informed prompts, we guide LLMs to extract personality-relevant information from text—the dominant modality in our study. To accommodate and amplify these personality-relevant representations, we further designed the Text-Centric Trait Fusion Network, a modality fusion architecture that treats text as the anchor while integrating auxiliary cues from audio and video modalities. Ablation studies show that our framework effectively enhances both the accuracy and stability of personality prediction. These advances provide a foundation for the development of systems that are more psychologically grounded, explainable, and adaptable in asynchronous video interviews (AVI) and beyond. In future work, we plan to explore adaptive prompt tuning and incorporate additional behavioral modalities to further enhance generalization and interpretability.

\begin{acks}
This work was supported by the National Natural Science Foundation of China (NSFC) under Grant Nos. 62202139 and 62172138. This work was also partially supported by the Fundamental Research Funds for the Central Universities under Grant Nos. JZ2025HGTB0226 and JZ2024HGTG0310. And the computation is completed on the HPC Platform of Hefei University of Technology.
\end{acks}
\bibliographystyle{ACM-Reference-Format}
\bibliography{sample-base}
\end{document}